\documentclass[letterpaper, 10 pt, conference]{ieeeconf}  

\IEEEoverridecommandlockouts                              

\usepackage{graphics} 
\usepackage{epsfig} 
\usepackage{mathptmx} 
\usepackage{times} 
\usepackage{amsmath} 
\usepackage{amssymb}  
\usepackage{bm}
\usepackage{graphicx}
\usepackage{siunitx}
\usepackage{nicefrac}
\renewcommand{\baselinestretch}{0.9}

\makeatletter
\let\NAT@parse\undefined
\makeatother

\usepackage[
  colorlinks=true,
  linkcolor=blue,
  urlcolor=blue,
  citecolor=blue
]{hyperref}
\title{\LARGE \bf
Optimal Control Approach for Non-prehensile Ball Juggling\\Using a 7-DoF Manipulator
}

\author{Joel Ramadani$^{1,*}$, Vasilije Rak\v{c}evi\'c$^{1}$, Riddhiman Laha$^{1}$, Arne Sachtler$^{1,3}$, Valentin Le Mesle$^{1}$,\\ Achim J. Lilienthal$^{1}$, and Sami Haddadin$^{2}$
\thanks{$^{*}$Corresponding author: \tt\small{joel.ramadani@tum.de}}
\thanks{$^{1}$Technical University of Munich, Munich, Germany.}
\thanks{$^{2}$Mohamed Bin Zayed University of Artificial Intelligence, Abu Dhabi, United Arab Emirates.}
\thanks{$^{3}$German Aerospace Center (DLR), Institute of Robotics and Mechatronics; Oberpfaffenhofen, Germany.}
}

\begin{document}

\maketitle
\thispagestyle{empty}
\pagestyle{empty}

\begin{abstract}
Non-prehensile object manipulation skills are important for real-world robot interactions, enabling highly dynamic tasks such as balancing a glass on a tray or the controlled sliding of items on a table. Among such tasks, those characterised by high-speed manipulation requirements and general sensitivity of the resulting hybrid dynamics are particularly hard to accomplish. Within these, juggling can be seen as a highly challenging maneuver to be solved. The key to robotic juggling is achieving dynamic stabilisation of an underactuated object. Since the object does not possess the ability of self-correction, its stability is entirely dependent on the forces applied to it. This creates a system that is sensitive to control inputs, where timing is critical to continuously counteract deviations and maintain the desired behavior. 
We develop a systematic method to control a 7-degree-of-freedom manipulator performing non-prehensile ball juggling with a tool. Our primary contribution is a model-based framework for generating juggling trajectories and stabilizing a periodic juggling motion for this hybrid system. The framework incorporates a two-stage optimal control approach to compute the underlying feasible motion patterns required for stable juggling. Offline-computed trajectories are then organised to enable real-time error correction without solving optimal control problems online. We demonstrate the effectiveness of the resulting controller by first evaluating its performance in a simulation environment and performing an experiment using a Franka Emika Panda robot.


\end{abstract}

\section{Introduction}

Robot non-prehensile manipulation has been widely studied \cite{ruggiero2018nonprehensile}, spanning tasks from quasi-static interactions to highly dynamic behaviors that impose significant hardware and control challenges~\cite{muchacho2022solution}. Robotic juggling represents one of the most demanding instances, with complexity strongly influenced by the manipulation strategy and physical setup.

One of the most commonly studied juggling tasks in literature is paddle juggling, where one or more balls are repeatedly hit into the air using a paddle (typically a racket or similar object). Here, the ball receives a brief impulse from the paddle during a short contact event that is modelled as an impulse exchange in \cite{buehler1987robotics}, \cite{aboaf1988task}. 
When modeled as a discrete nonlinear system, rhythmic paddle juggling can exhibit inherent self-stabilizing properties that allow open-loop execution \cite{schaal1996one, schaal1993open, reist2012design}.
However, longer contact duration in juggling based on non-prehensile object manipulation invalidates models based on purely impulsive events. Additionally, the self-stabilising property of these systems is not necessarily inherent for all juggling tasks.

\begin{figure}[t]
\centering
\includegraphics[scale=0.335]{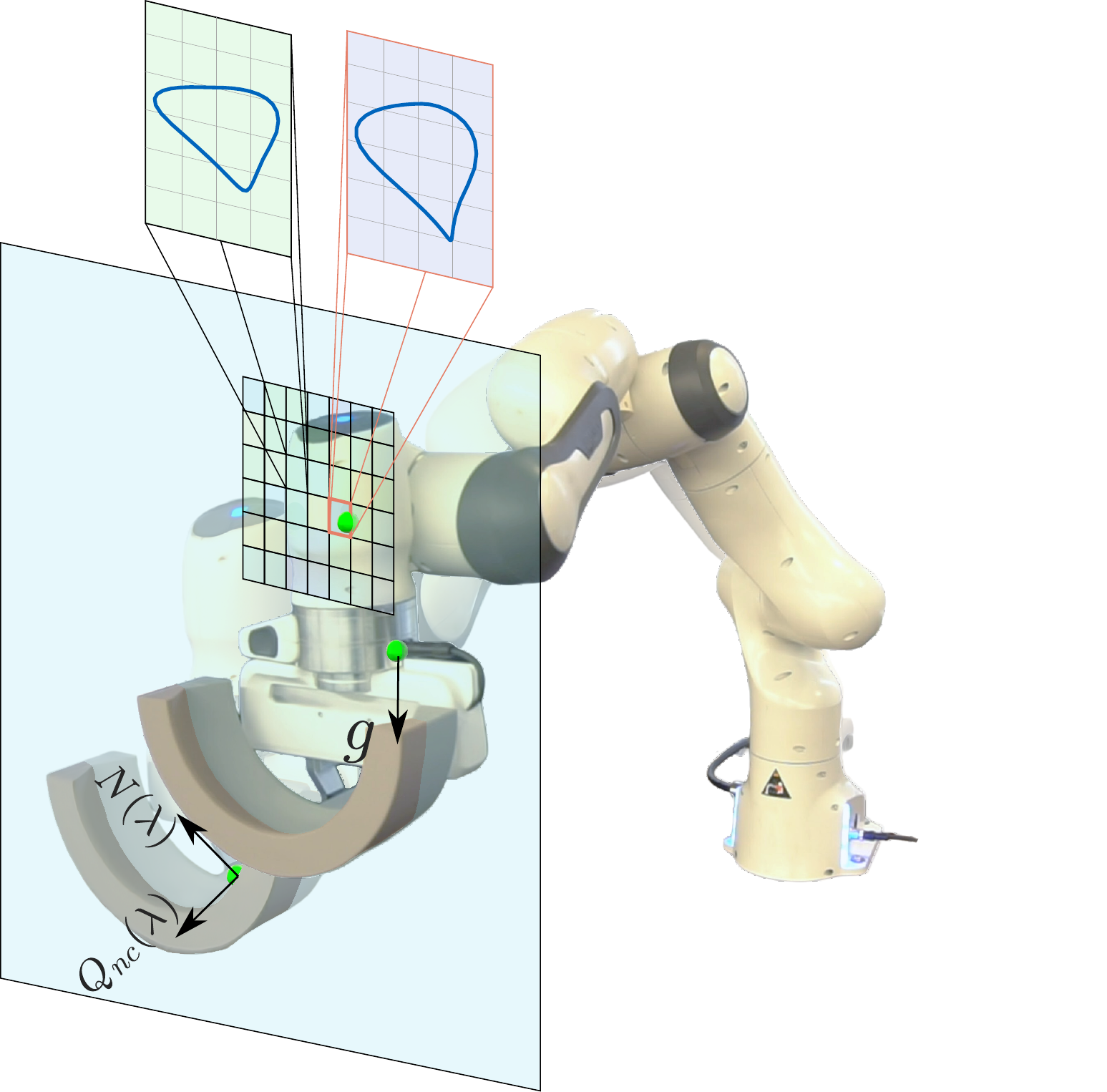}
\caption{Visualized juggling setup with a representation of the juggling plane and decision-making policy. Non-conservative force $Q_{nc}$ and Normal force $N$ are modelled in relation to the Lagrange multiplier $\lambda$. For the non-contact (flight) phase, ball motion is only influenced by gravity.}
\label{fig:poster}
\vspace{-0.5cm}
\end{figure}

In contrast to paddle juggling, toss juggling requires the ball to stay in contact for a longer duration of time. In robotics, funnel-shaped tools attached to the end-effector of a manipulator are often used to catch the ball. An analytical framework for modelling the task is introduced in \cite{beek1995science, polster2003mathematics}. 
The difficulty of catching strongly depends on the mechanical design, as deep funnel-shaped tools can passively prevent bouncing and reduce the task to trajectory interception.
Various trajectory optimization strategies have been proposed to improve catching performance, including energy-efficient motion planning \cite{lampariello2011trajectory, bauml2010kinematically} and acceleration matching to reduce sensitivity to impact uncertainties \cite{hove1991experiments, namiki2014ball}. Learning-based approaches have further been explored to enhance robustness in complex catching scenarios \cite{kim2014catching}. These approaches primarily focus on accurate interception and impact handling, assuming that the ball is passively retained after contact (caught). In contrast, our task involves prolonged contact with complex dynamics, where the ball must be actively guided.



Beyond catch-toss juggling forms, relevant approaches also include scenarios of a ball that is continuously balanced or guided along surfaces without leaving contact. 
The first-order kinematic equations describing the contact between two three-dimensional objects are derived in \cite{montana1988kinematics}. An optimization framework for planning and controlling first-order rolling interactions between rigid bodies is described in \cite{woodruff2020motion}. Modelling interaction between the tool and object, alongside active control throughout the entire contact phase to guide the ball to a suitable launch state is at the core of the tackled problem. However, it's strongly coupled to the flight phase dynamics and thus can not be considered in separation. 


Finally, planning methods used in contact juggling \cite{woodruff2023robotic} are ill-suited for the highly dynamic nature of our task, which requires fast decision cycles and explosive motions to maintain stability.
Consequently, real-time trajectory adaptation via online optimization becomes impractical, as solving the underlying Optimal Control Problem (OCP) exceeds the millisecond-scale time budget imposed by the task dynamics.
Despite extensive work on impulsive paddle juggling, funnel-based catching, and continuous rolling interactions, control of prolonged-contact tool-based juggling remains largely unexplored.

This motivates the development of a new control architecture that can represent the task's complex hybrid dynamics adapted for real-time, high-speed motion generation.

\section{Overview}

Our main contribution is a model-based framework for feasible trajectory generation and sustained juggling for non-prehensile ball manipulation. Juggling involves a swing-up phase for launching the ball in the air, followed by controlled juggles as illustrated in Fig.~\ref{fig:juggling_task}. All aspects of the framework can be followed through Fig.~\ref{fig:juggling_approach}. We begin by detailing the modelling and hybrid dynamics definition for the tool-ball interaction in Section~\ref{sec:Dynamics_Model}. Further, a two-stage OCP is introduced in sections \ref{sec:TS_OCP} and \ref{sec:JS_OCP} that results in feasible swing-up and juggling trajectories that can be executed on the robot. A scheme for an online error correction via a database of trajectories is introduced in \ref{sec:online_stabilization}. Simulation validation results and robot hardware experiment are presented in \ref{sec:experiments}.  

\subsection{Ball Juggling Task} \label{sec:Juggling_Definition}

\begin{figure}[t]
\vspace{0.17cm}
\centering
\includegraphics[scale=0.33]{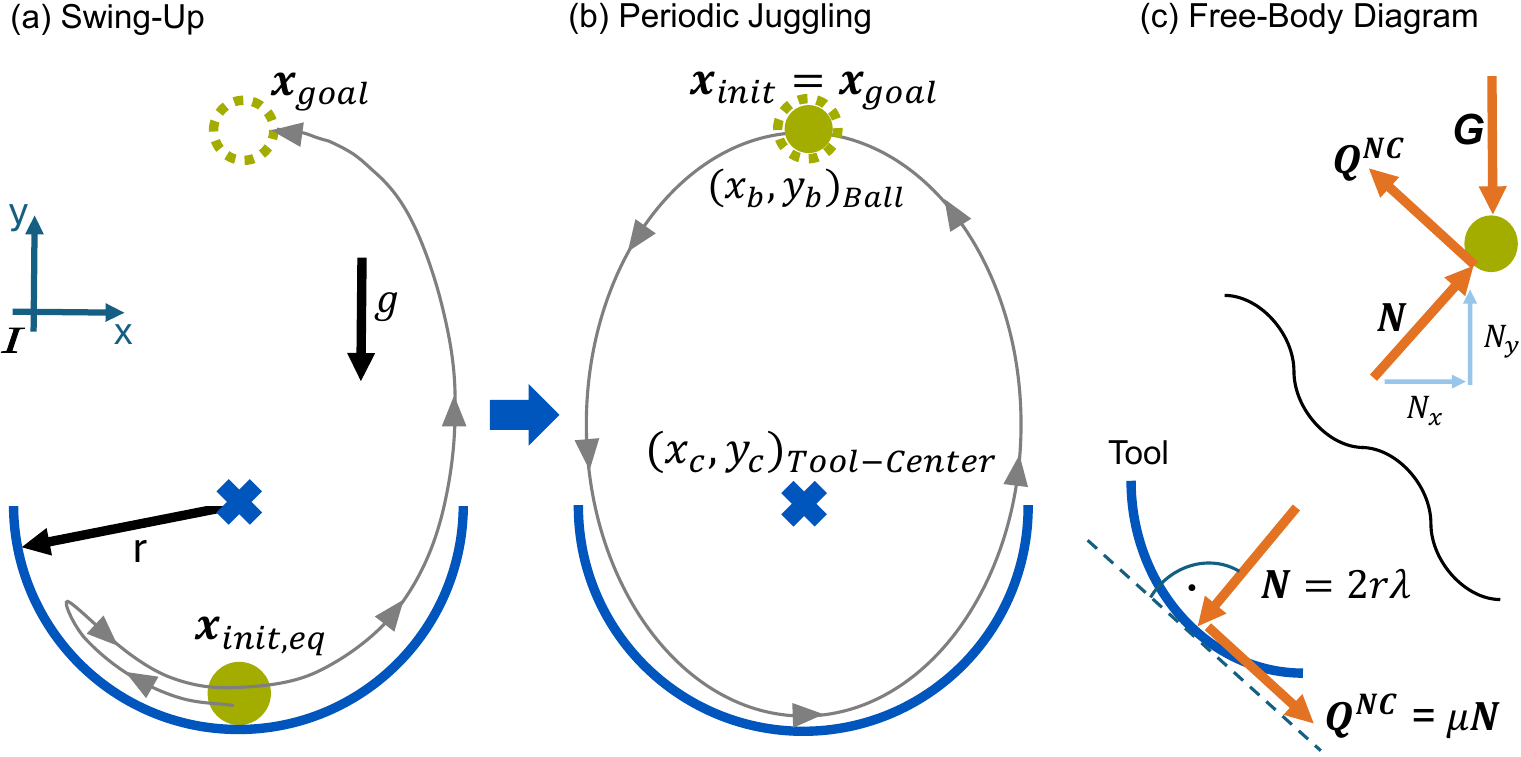}
\caption{Planar representation of the juggling task. (a) Swing Up phase; (b) Periodic juggling phase; (c) Free-Body Diagram of the Tool-Ball System}
\label{fig:juggling_task}
\vspace{-0.5cm}
\end{figure}

Non-prehensile ball manipulation for juggling is performed via a tool (a half-circular tool with a V groove) mounted to the end-effector of a robotic manipulator. Initially, the ball is at rest in a stable equilibrium position of the tool $\bm{x}_{\mathrm{init,eq}}$. In the first phase, referred to as the swing-up (Fig.~\ref{fig:juggling_task} (a)), the robot aims to launch the ball into the air and bring it into a desired dynamic state, characterized by a target position and velocity. In particular, within this phase, the tool is actuated so that the ball gains enough energy to exit the tool at its edge. Subsequently, the ball loses contact and enters a flying trajectory, where its motion is only affected by gravity. Once the desired system state $\bm{x}_{\mathrm{goal}}$ is reached, the objective is to maintain continuous juggling behaviour (Fig.~\ref{fig:juggling_task} (b)): as the ball starts falling after reaching its apex, it is caught by the tool and guided back into the desired state by coordinated movements of the tool. The target state is defined by the ball's apex state vector $\bm{x}_{\mathrm{goal}} = [x_b^*, y_b^*, \dot{x}_b^*, \dot{y}_b^*]^{\mathrm{T}}$, where the subscript $b$ indicates the ball and the asterisk ($*$) denotes the desired value for each state component. This state defines the desired apex of the flight parabola, corresponding to a specific juggling height $y_b^*$ and a horizontal ball velocity $\dot{x}_b^*$. By definition of the apex, the vertical velocity $\dot{y}_b^*$ is zero.

\subsection{Approach} \label{sec:Approach}

\begin{figure}[t]
\vspace{0.17cm}
\centering
\includegraphics[scale=0.37]{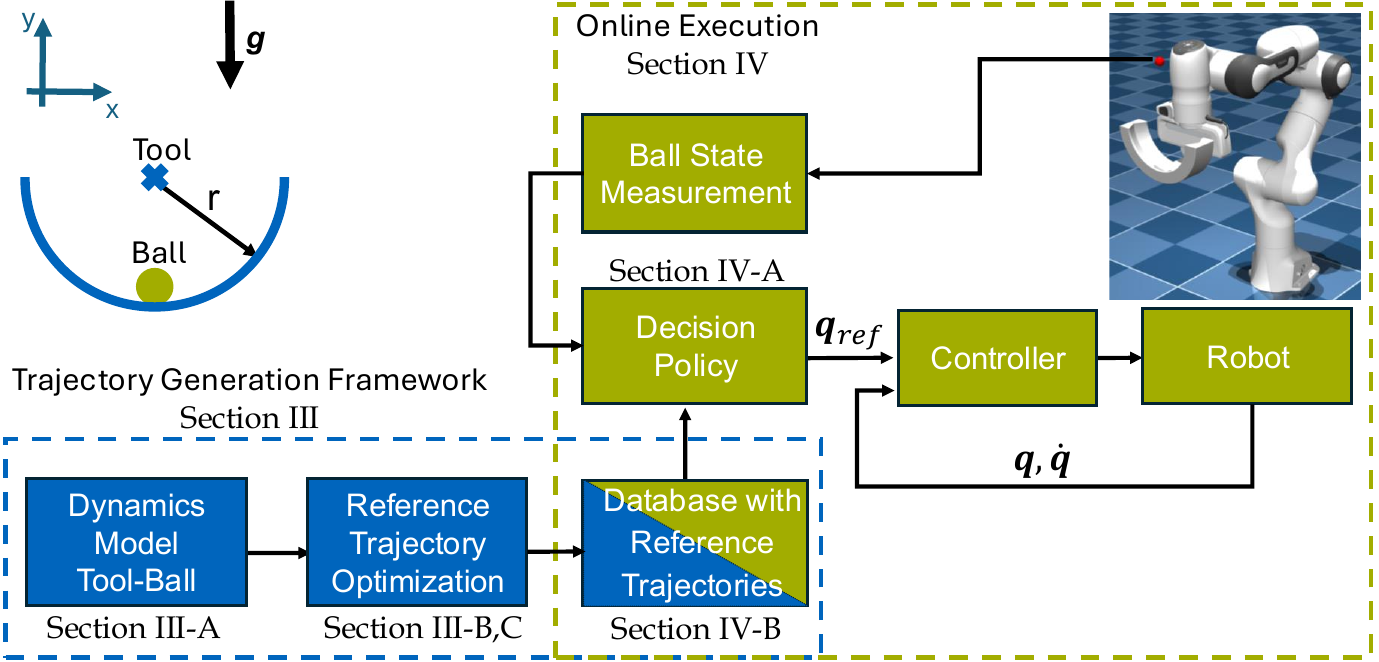}
\caption{Block Diagram of the Juggling Control Framework}
\label{fig:juggling_approach}
\vspace{-0.5cm}
\end{figure}

The developed control framework for stable juggling, as illustrated in Fig.~\ref{fig:juggling_approach}, relies on a library of pre-computed optimal trajectories for fast online trajectory adaptation. Each trajectory is generated offline in a two-stage process. First, an OCP founded on a two-dimensional model of the tool-ball interaction, computes kinematically ideal motions. Subsequently, a second OCP maps these task-space paths into dynamically feasible joint-space trajectories, ensuring that all of the robot's physical constraints are respected.

To ensure robustness against model inaccuracies and external disturbances, the two-stage optimization process is repeatedly solved offline for various boundary conditions and generates a trajectory database. The database populates the state space around the nominal target apex \(\bm{x}_{\mathrm{goal}}\), providing a library of recovery motions for a variety of perturbed ball states.

In the online phase, the ball's state is measured as it leaves the tool, and its subsequent flight path is estimated via flight dynamics. Based on this estimation, a decision policy selects the most suitable recovery trajectory from the database. The robot's joint-level velocity controller is used for the trajectory tracking. The overall result is a continuous and stable juggling pattern.

\section{Trajectory Generation Framework}\label{sec:trajectory_generation}

For the generation of the dynamically feasible juggling trajectories, the problem is decoupled into two main stages, each formulated as a distinct OCP:

\begin{enumerate}
    \item \emph{Stage 1. Task-Space OCP} – generates an ideal reference trajectory for the tool's motion based on a 2D reference model. The model captures the physics of the ball-tool interaction.  
    \item \emph{Stage 2. Joint-Space OCP} – takes the 2D tool trajectory and maps it onto the full robot model. The objective here is to find a complete, dynamically feasible joint-space motion that accurately tracks the path from the first stage.
\end{enumerate}

At the core of the OCP lies a computationally efficient formulation of the system dynamics model.

\subsection{Dynamics Model} \label{sec:Dynamics_Model}
We begin the analysis by examining the fundamental dynamics of the system: the motion of a ball within a freely movable tool. 

\subsubsection{Assumptions} 

To ensure mathematical tractability while preserving key physical properties, we introduce the following assumptions:

\begin{itemize}
    \item \emph{Planar Point-Mass System:} Both tool and ball are constrained to move within the vertical xy-plane. The ball is modelled as a point mass, neglecting rotational dynamics.

    \item \emph{Mass-less Kinematic Tool:} The tool's dynamics (inertia) are ignored; it serves as an ideal kinematic driver for the ball with a fixed orientation.

    \item \emph{Unilateral Interaction:} The reaction force from the ball onto the tool is considered negligible due to the small mass ratio (Ball $\ll$ Robot).
\end{itemize}

These assumptions enable the formulation of a reduced model focusing on the ball's interaction dynamics in the first stage, and on the robot dynamics in the second. Consequently, the impact from the ball is treated as an external disturbance that is compensated by the feedback controller driving the robot.

\subsubsection{Equations of Motion}

Let \(I^X, I^Y\) be the inertial Cartesian coordinate system presented in Fig.~\ref{fig:juggling_task} (a), and \((x_c, y_c)\), \((x_b, y_b)\) be the coordinates of the tool-center and the ball, respectively.

To describe the motion of the ball while in contact with the tool, the Lagrangian formalism \cite{lagrange1853mecanique} is used, extended by a Lagrange multiplier $\lambda$ to enforce the physical constraint:
\begin{equation}
	L = T - U + \lambda C
	\label{eq:lagrangian_general}
\end{equation}
where $T$ and $U$ are the kinetic and potential energy of the ball, respectively, and the term $\lambda C$ enforces the geometric constraint that the ball remains on the tool's surface:

\begin{equation}
	r^2=(x_b - x_c)^2 + (y_b - y_c)^2
	\label{eq:lagrange_c_constraint}
\end{equation}

With the ball's mass $m$ and the tool's radius $r$, the specific Lagrangian for the system is given by:

\vspace{-0.3cm} 
\begin{equation}
L = \underbrace{\frac{1}{2}m\left(\dot{x}_b^2 + \dot{y}_b^2\right)}_{T} - \underbrace{mg y_b \vphantom{\frac{1}{2}}}_{U} + \lambda \underbrace{\left((x_b - x_c)^2 + (y_b - y_c)^2 - r^2\right) \vphantom{\frac{1}{2}}}_{C}
\label{eq:lagrangian}
\end{equation}

By defining the vector of generalized coordinates as $\boldsymbol{\eta} = [x_b, y_b]^T$, the Euler-Lagrange equations take the general form:

\begin{equation}
	\frac{d}{dt} \frac{\partial L}{\partial \dot{\boldsymbol{\eta}}} - \frac{\partial L}{\partial \boldsymbol{\eta}} = \boldsymbol{Q}_{\text{nc}}
	\label{eq:euler_lagrange}
\end{equation}
where the vector of non-conservative forces, given by $\boldsymbol{Q}_{\text{nc}} = [Q_x^{\text{nc}}, Q_y^{\text{nc}}]^T$, is used to model Coulomb friction in the system. Evaluating the partial derivatives of the Lagrangian from Equation~\eqref{eq:euler_lagrange} yields the following equations of motion:

\begin{subequations}\label{eq:eom}
\vspace{-0.3cm}
\begin{align}
    m \ddot{x}_b 
    - \underbrace{2\lambda(x_b - x_c)}_{N_x} 
    &= Q_x^{\text{nc}} \label{eq:eom_x} \\
    m \ddot{y}_b
    + mg
    - \underbrace{2\lambda(y_b - y_c)}_{N_y} 
    &= Q_y^{\text{nc}} \label{eq:eom_y}
\end{align}
\end{subequations}
where the non-conservative forces $Q_x^{\text{nc}}$ and $Q_y^{\text{nc}}$ model the Coulomb friction. They are defined as:
\begin{subequations}\label{eq:Q_lambda}
\begin{align}
	Q_x^{\text{nc}} &= -\mu \cdot 2 \lambda r \cdot \hat{T}_x \cdot \text{sgn}\big(\langle \boldsymbol{v}_{\text{rel}}, \boldsymbol{\hat{T}} \rangle \big) \label{eq:Qxnc_lambda} \\
	Q_y^{\text{nc}} &= -\mu \cdot 2 \lambda r \cdot \hat{T}_y \cdot \text{sgn}\big(\langle \boldsymbol{v}_{\text{rel}}, \boldsymbol{\hat{T}} \rangle \big) \label{eq:Qync_lambda}
\end{align}
\end{subequations}

where $\mu$ is the friction coefficient, $\boldsymbol{\hat{T}}$ is the unit tangent vector at the contact point, and $\boldsymbol{v}_{\text{rel}}$ is the relative velocity between the ball and the tool:
\begin{equation}
	\boldsymbol{\hat{T}} = \frac{1}{r} \begin{pmatrix} y_b - y_c \\ -(x_b - x_c) \end{pmatrix},
	\qquad
	\boldsymbol{v}_{\text{rel}} = \begin{pmatrix} \dot{x}_b - \dot{x}_c \\ \dot{y}_b - \dot{y}_c \end{pmatrix}
	\label{eq:kinematic_defs}
\end{equation}
The signum function on the dot product, $\text{sgn}(\langle \boldsymbol{v}_{\text{rel}}, \boldsymbol{\hat{T}} \rangle)$, ensures that the friction force always opposes the tangential component of the relative motion. Note that~\eqref{eq:eom_x} and~\eqref{eq:eom_y} govern the dynamics only as long as the ball is in contact with the tool. For completeness, the dynamics of the ball during free fall can be described by the following equations:
\begin{equation}
	\ddot{x}_b = 0, \qquad \ddot{y}_b = -g.
	\label{eq:free_fall}
\end{equation}

Examining~\eqref{eq:eom_x} and~\eqref{eq:eom_y} and comparing them to the free fall dynamics in~\eqref{eq:free_fall}, the choice of the Cartesian over the polar coordinate system becomes clear. Substituting $\lambda=0$ in~\eqref{eq:eom_x} and~\eqref{eq:eom_y} yields equations that coincide with the free fall dynamics in~\eqref{eq:free_fall}. The presented equations result in a uniform description of the system's hybrid dynamics.

\subsubsection{The Lagrange Multiplier}

The components $N_x = 2\lambda(x_b - x_c)$ and $N_y = 2\lambda(y_b - y_c)$ in~\eqref{eq:eom_x} and~\eqref{eq:eom_y} define a constraint force vector $\boldsymbol{N} = [N_x, N_y]^T$
(see Fig.~\ref{fig:juggling_task} (c)). The magnitude of the normal force $N$ exerted by the tool is given by $N = \sqrt{N_x^2 + N_y^2}$. This expression simplifies to $N = 2\lambda r$. Consequently, the Lagrange multiplier $\lambda$ can be physically interpreted as the normal force scaled by the tool's diameter, i.e., $\lambda = \nicefrac{N}{2r}$.

The value of $\lambda$ can be determined using the holonomic constraint  for keeping the ball movement on the tool's surface, defined in~\eqref{eq:lagrange_c_constraint}. By differentiating this constraint equation twice with respect to time and substituting the equations of motion for $\ddot{x}_b$ and $\ddot{y}_b$ (from~\eqref{eq:eom_x} and~\eqref{eq:eom_y} without the contribution of the friction forces), one can solve for $\lambda$:
\begin{equation}
	\lambda = \frac{m}{2r^2} \left[ -\|\boldsymbol{v}_{\text{rel}}\|^2 + g(y_b - y_c) + \ddot{x}_c(x_b - x_c) + \ddot{y}_c(y_b - y_c) \right]
	\label{eq:lambda_scalar}
\end{equation}

An interesting property of derived dynamics~\eqref{eq:eom} is that a closed-form solution for components of the normal force $N$, and subsequently $\lambda$ from~\eqref{eq:lambda_scalar}, is not required to be included in the dynamics. $\lambda$ is rather defined as an optimization parameter that abstracts the interaction between the tool and a ball. Unidirectional physical property is encoded via constraint $\lambda(t) \geq 0$.

\subsection{Task-Space Optimal Control for Tool Trajectory Generation} \label{sec:TS_OCP}

Based on the derived system dynamics, we formulate a \emph{{Task-Space OCP}} to generate tool motions for both the swing-up and periodic juggling of the ball. The control input $\boldsymbol{u}(t)$ is the tool's Cartesian acceleration, $\boldsymbol{u}(t) = [\ddot{x}_c(t), \ddot{y}_c(t)]^T$, while the state $\boldsymbol{x}(t)$ comprises the positions and velocities of the ball and tool.

A key challenge is the system's hybrid nature, alternating between contact and free fall. We address this by formulating the OCP as a Mathematical Program with Complementarity Constraints (MPCC), such that the transition is managed by a complementarity condition subject to frictionless contact constraints \cite{posa2014direct,yao2025synthesis}. The Lagrange multiplier $\lambda$ is introduced as a decision variable for the complementarity condition in the optimization. This condition is parametrized by a smooth signed distance function $\phi(\boldsymbol{x}, \alpha)$, representing log-sum-exp with a smoothing parameter $\alpha$. It reaches zero when the ball is in contact with the tool's surface.

\vspace{-0.3cm} 
\begin{equation}
\phi(\boldsymbol{x}, \alpha) =
\frac{1}{\alpha} \log\!\left[
 e^{\alpha(r^2-(x_b-x_c)^2-(y_b-y_c)^2)}
 + e^{\alpha(y_b-y_c)}
\right]
\end{equation}

With this, the complete OCP, which finds the optimal trajectories and final time $t_f$, is formulated as:
\begin{subequations}\label{eq:ocp_mpcc_formulation}
\begin{align}
\min_{\boldsymbol{x}(\cdot),\,\boldsymbol{u}(\cdot),\,\lambda(\cdot),\,t_{f}} \quad & J = \int_0^{t_f} \| \boldsymbol{u}(t) \|^2\, \mathrm{d}t \label{eq:ocp_cost_mpcc}\\
\text{such that} \quad & \dot{\boldsymbol{x}}(t) = f(\boldsymbol{x}(t), \boldsymbol{u}(t), \lambda(t)), \quad t \in [0, t_{f}] \label{eq:ocp_dynamics_mpcc}\\
& b(\boldsymbol{x}(0), \boldsymbol{x}(t_{f})) = 0 \label{eq:ocp_boundary_constraints_mpcc}\\
& 0 \le \lambda(t) \perp \phi(\boldsymbol{x}(t), \alpha) \ge 0, \; t \in [0, t_{f}] \label{eq:ocp_comp_condition}
\end{align}
\end{subequations}
The objective~\eqref{eq:ocp_cost_mpcc} minimizes the control effort. While higher-order terms like jerk penalties were evaluated, this formulation yielded improved convergence behavior and smoother acceleration profiles better suited for tracking on the current hardware. The Problem is subject to the system dynamics~\eqref{eq:ocp_dynamics_mpcc} and boundary conditions~\eqref{eq:ocp_boundary_constraints_mpcc}. The core of the hybrid model lies in the complementarity condition~\eqref{eq:ocp_comp_condition}, which enforces the contact logic.

\subsubsection{Solving the \emph{Task-Space OCP}}
The OCP is formulated and solved using CasADi \cite{andersson2019casadi} within MATLAB. It is transcribed into a nonlinear program (NLP) using a direct collocation method with Gauss-Legendre integration points. The NLP is solved using the interior point solver IPOPT \cite{wachter2006implementation} in combination with the linear solver MUMPS. Additionally, the linear solver MA97 from the HSL \cite{hsl2007collection} library was evaluated, and both combinations produced comparable results. However, MA97 showed improved computational speed. 

We solve the \textit{\emph{Task-Space OCP}}~\eqref{eq:ocp_mpcc_formulation} separately for the swing-up motion and the periodic juggling motion, since the transition point is predefined by the desired ball apex state. The main difference in formulations lies in the boundary conditions~\eqref{eq:ocp_boundary_constraints_mpcc}: For the swing-up, the ball starts in the equilibrium position $\bm{x}_{\mathrm{init,eq}}$ and reaches $\bm{x}_{\mathrm{goal}}$ as terminal boundary condition (cf. Fig.~\ref{fig:juggling_task}a), whereas for the periodic juggling task, $\bm{x}_{\mathrm{goal}}$ is imposed as boundary condition at both the beginning and the end of the trajectory (cf. Fig.~\ref{fig:juggling_task}b).

No path constraints are imposed on the tool’s velocity or acceleration. Instead, only the Lagrange multiplier $\lambda(t)$ is bounded, which implicitly constrains the tool accelerations through the ball-tool interaction (cf.~\eqref{eq:lambda_scalar}). This choice showed an improved numerical convergence behaviour and also allows enforcing smoother catches as an upper bound on $\lambda$ effectively limits the initial impact force when the ball is caught by the tool.

\begin{figure*}[t]
\vspace{0.17cm}
\centering
\includegraphics[scale=0.67]{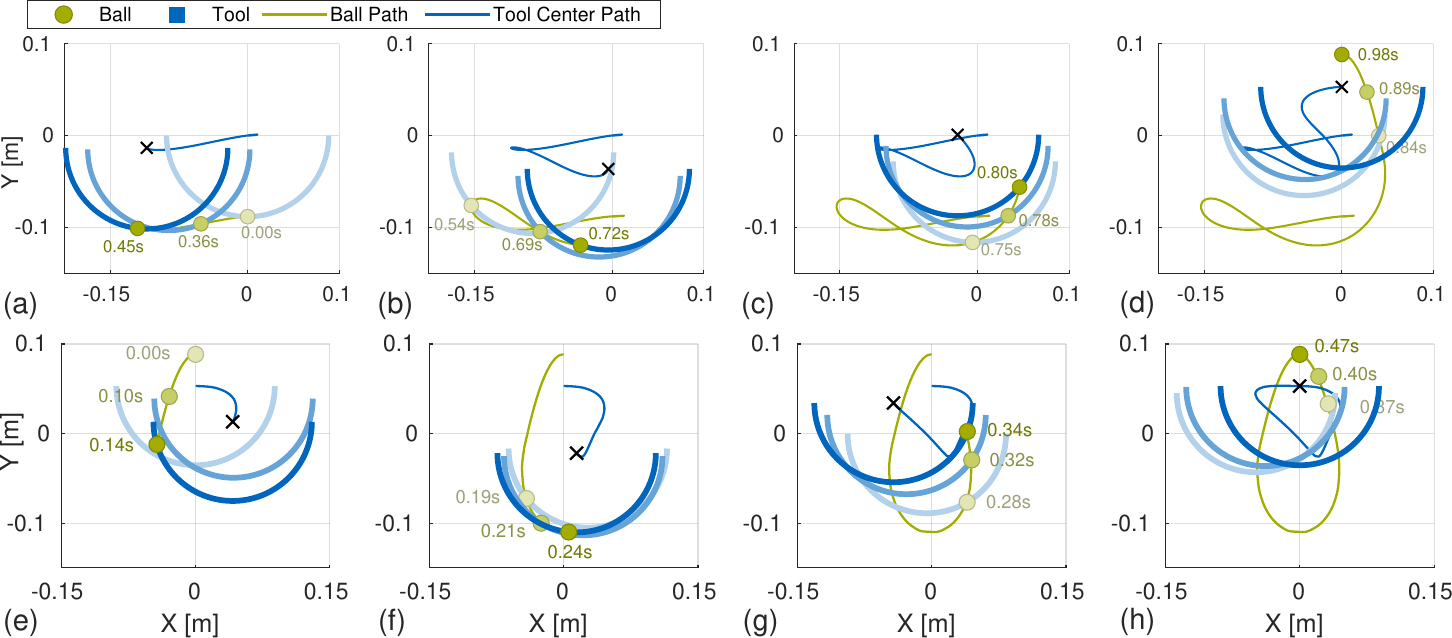}
\caption{Snapshots of a reference trajectory obtained from the \emph{{Task-Space OCP}}. (a-d) Swing-up motion of the tool and ball; (e-h) Periodic juggling cycle.}
\label{fig:full_juggling_task}
\vspace{-0.5cm}
\end{figure*}

\subsection{\emph{Joint-Space OCP} for Feasible Reference Trajectory Tracking} \label{sec:JS_OCP}

Since the task requires the robot arm for tool manipulation, its limits have to be taken into account. For transferring generated 2D task-space trajectories to a Franka Emika Panda~\cite{haddadin2022franka} robot, a standard Jacobian-based Cartesian controller is not sufficient, because the trajectories for both the swing-up and the juggling are highly dynamic. Additionally, kinematic constraints require the tool to stay on a 2D task space, which can introduce additional difficulties for Cartesian speed tracking. Thus, to translate the planar reference trajectory into physically feasible joint trajectories for the 7-DoF robot, a second OCP~\eqref{eq:ocp_robot_mapping} was defined to incorporate the full system dynamics and actuation constraints.

The process begins by preprocessing the trajectories from the initial 2D juggling model. The time intervals where the ball is in contact with the tool are identified by using the Lagrange multipliers associated with the contact constraints of the task-space OCP. This is done because precise trajectory tracking is only required while the ball is in contact with the tool. The adopted 2D tool path is then embedded in the robot's 3D workspace as  $\bm{p}_{des}$. To find a suitable workspace region characterised by high agility, a manipulability analysis was performed on the Franka Panda. The forward kinematics of this configuration define the initial end-effector pose, to which the 2D trajectory is rigidly attached. 

The \emph{{Joint-Space OCP}} is formulated to find a feasible joint-space trajectory that follows the desired tool trajectories, while simultaneously maintaining the tool in a desired orientation.

\begin{subequations}\label{eq:ocp_robot_mapping}
\begin{align}
    \text{find} \quad & \bm{x_r}(\cdot),\,\bm{u_r}(\cdot) \nonumber \\
    \text{subject to} \quad & \dot{\bm{x}}_r(t) = f(\bm{x_r}(t), \bm{u_r}(t)), \quad t \in [t_\text{init}, t_\text{goal}] \label{eq:ocp_map_dynamics} \\
    & \bm{q}_{\min} \le \bm{q}(t) \le \bm{q}_{\max} \label{eq:ocp_map_q_limits} \\
    & |\bm{u}_{r} (t)| \le 0.9 \ \bm{u}_{r,\max} \label{eq:ocp_map_u_limits} \\
    & |\dot{\bm{u}}{r} (t)| \le 0.9 \ \dot{\bm{u}}_{r,\max} \label{eq:ocp_map_du_limits} \\ 
    & \bm{p}_{\mathrm{ee}}(\bm{q}(t)) = \bm{p}_{\mathrm{des}}(t), \quad \forall t \in T_c \label{eq:ocp_map_pos_constr} \\
    & \bm{R}_{\mathrm{ee}}(\bm{q}(t)) = \bm{R}_{\mathrm{des}}, \quad \forall t \in T_c \label{eq:ocp_map_ori_constr} \\
    & \bm{x_r}(t_\text{init}) = \bm{x_r}(t_\text{goal}) \label{eq:ocp_map_cyclic_constr}
\end{align}
\end{subequations}

Here, $\bm{x_r} = (\bm{q}, \dot{\bm{q}})$ represents the robot's joint states. The path constraints for joint torques $\bm{u}_r$~\eqref{eq:ocp_map_u_limits} and their rate of change $\dot{\bm{u}}_{r}$~\eqref{eq:ocp_map_du_limits} were chosen, as these are the most versatile limits specified by the Franka Control Interface (FCI). The torque limits in the OCP were set to 90\% of the robot's maximum capabilities to ensure a feedback controller can react to tracking errors and external disturbances. 

The dynamics constraint \eqref{eq:ocp_map_dynamics} encodes the standard robot equations of motion:
\begin{equation}
\dot{\bm{x}}_r =
\begin{bmatrix}
\dot{\bm{q}} \\
M(\bm{q})^{-1} \big( \bm{u}_r - C(\bm{q}, \dot{\bm{q}})\dot{\bm{q}} - g(\bm{q}) \big)
\end{bmatrix},
\label{eq:robot_dynamics}
\end{equation}
where $M(\bm{q})$ is the joint-space mass matrix, $C(\bm{q}, \dot{\bm{q}})$ contains the Coriolis and centrifugal terms, and $g(\bm{q})$ represents gravity. We constrain the forward kinematics of the robot to track the desired tool trajectory by enforcing the position and orientation constraints \eqref{eq:ocp_map_pos_constr}–\eqref{eq:ocp_map_ori_constr} only during the contact phases $T_c$. This gap leaves the initial and final segments of the trajectory unconstrained and provides the optimizer with more flexibility to plan smooth transitions between juggling cycles. 

The \emph{{Joint-Space OCP}}~\eqref{eq:ocp_robot_mapping} is again formulated as a NLP using direct collocation and the CasADi~\cite{andersson2019casadi} framework. To improve solving speed, the Pinocchio~\cite{carpentier2019pinocchio} library was adopted, as it provides faster differentiation of the complete robot dynamics and its integration with CasADi ensures highly efficient symbolic processing. 


\section{Stabilization of the Juggling Motion} \label{sec:online_stabilization}
\subsection{Decision Policy}
A Poincar\'e analysis~\cite{Ramadani2025} confirms that the optimized trajectories do not exhibit self-stabilizing properties under open-loop execution, reflecting the inherent instability of the system. Thus, they are not sufficient for sustained juggling: the ball tends to fall out of the tool after a few cycles when executing the same tool trajectory periodically. To address this, we introduce a feedback mechanism that adapts the tool's motion once per cycle based on the observed ball state.

For deciding on a trajectory for the next cycle, a ball state at  $t_{\mathrm{release}}$ is measured. The ball's position and velocity $\widetilde{\bm{x}}_b$ are used to predict its future state at the apex $\hat{\bm{x}}_b$ of its parabolic flight path (see Fig.~\ref{fig:decision_policy}). This predicted apex state is then matched to the closest entry in the pre-computed trajectory database, using the minimum Euclidean distance \( d_{\min} \), and the corresponding tool trajectory is selected for the upcoming cycle. By estimating the apex state immediately at release, the system can utilize the flight phase to execute a smooth, continuous transition to the newly selected trajectory before the ball needs to be caught again. 

\begin{figure}[t]
\vspace{0.16cm}
\centering
\includegraphics[scale=0.31]{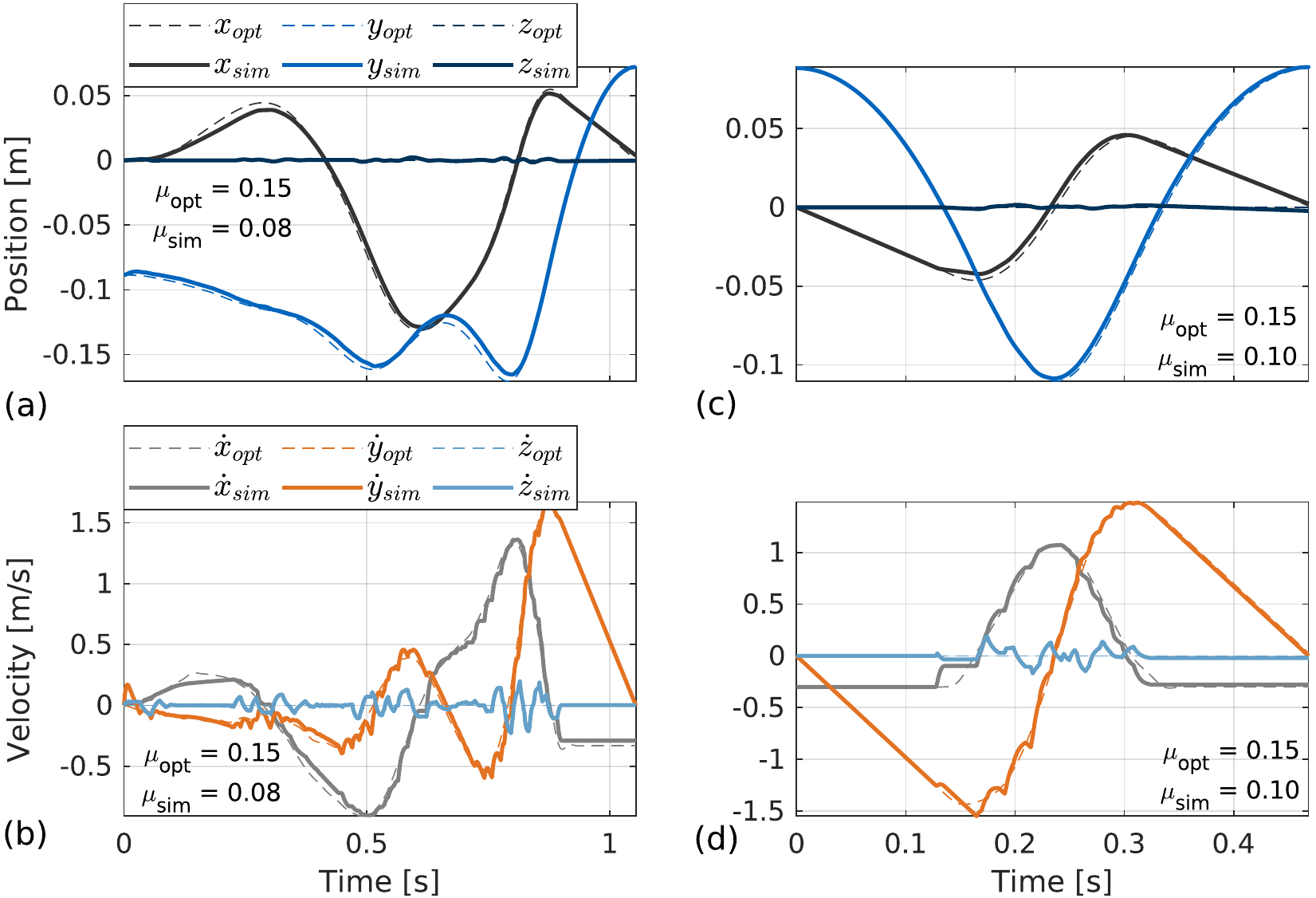}
\caption{Comparison of the optimal and simulated ball trajectories for the swing-up (a-b) and juggling (c-d) maneuvers.}
\label{fig:Opt_vs_Sim}
\vspace{-0.5cm}
\end{figure}

\subsection{Database generation}
\begin{table}[b]
    \centering
    \vspace{-0.5cm}
    \caption{Model Parameters for Trajectory Database}
    \label{tab:ocp_params}
    \begin{tabular}{l c S[table-format=1.5]}
        \hline
        \textbf{Parameter} & \textbf{Symbol} & {\textbf{Value}} \\
        \hline
        Gravitational acceleration & $g$ & \SI{9.81}{\meter\per\second\squared} \\
        Ball mass & $m$ & \SI{0.1}{\kilogram} \\
        Tool radius & $r$ & \SI{0.08845}{\meter} \\
        Tangential friction coeff. & $\mu$ & 0.17 \\
        \hline
    \end{tabular}
\end{table}

\begin{figure}[t]
\vspace*{0.2cm}
\centering
\includegraphics[scale=0.4]{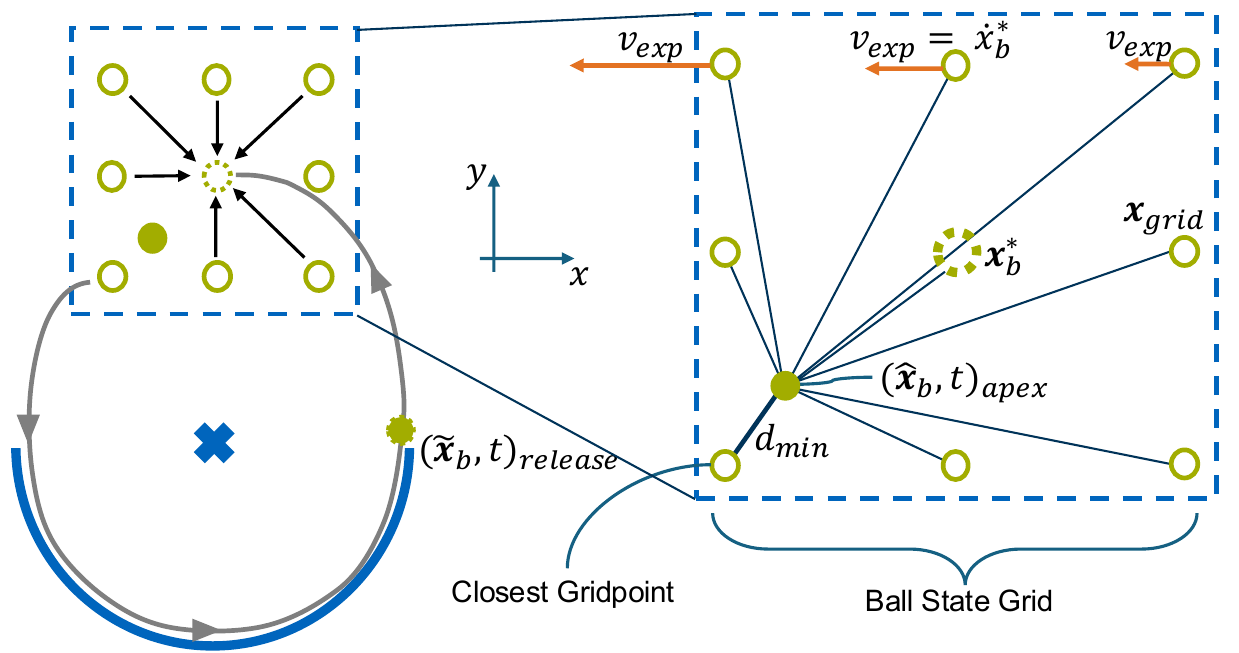}
\caption{Representation of the trajectory database combined with the decision policy to select the next trajectory to perform. The ball state is measured at release, and its apex state is estimated. Based on this estimation, a suitable trajectory is retrieved from the database.}
\label{fig:decision_policy}
\vspace{-0.5cm}
\end{figure}

To generate the database, the \emph{{Task-Space OCP}} is solved repeatedly in a loop. Since the goal is to correct ball trajectory deviation by bringing the ball back to the correct apex (see Fig.~\ref{fig:decision_policy}), the database contains trajectories associated with all perturbed apex states of the ball (a square grid of apex positions spanning \(5\text{ cm} \times 5\text{ cm}\) around the desired apex, with a resolution of \(3.125\text{ mm}\)) as the initial boundary condition. The terminal boundary is fixed at the desired apex. For each grid point, several candidate horizontal velocities are assigned. The expected velocity \( v_{\exp} \) at the apex is estimated as a linear function of the horizontal offset, such that balls reaching their apex left of the desired point are associated with higher horizontal velocities than those reaching it on the right (see Fig.~\ref{fig:decision_policy}). This ensures that the database captures the velocity error trend for counterclockwise ball motion. Indeed, for the fixed release position, the balls that end up further left have higher horizontal velocity at the apex.

Each consecutive solution (ball and tool trajectories) takes the previous iteration as a warm start for the current problem. This approach aims to prevent the solutions from drifting apart over iterations. By drifting, we mean that, due to the non-convexity of the problem, multiple distinct solutions exist that successfully return the ball to the desired state. For instance, it was observed that some solutions resulted in the ball being caught with prolonged contact times, differing significantly from the solution presented in Fig.~\ref{fig:full_juggling_task}b.

To map each of these reference trajectories to the full robot model and ensure their dynamic feasibility, \emph{{Joint-Space OCP}} is utilized. First, a single trajectory with only a periodic boundary condition, $\bm{x}_{r,\mathrm{init}} = \bm{x}_{r,\mathrm{goal}}$ is solved. Its solution establishes a feasible common reference state $\bm{x}_{r,\mathrm{apex}}$ at $t_{\mathrm{apex}}$ for the robot. For all subsequent trajectories in the database \emph{{Joint-Space OCP}}s are then solved with modified boundary constraints to match the initially found reference state: $\bm{x}_{r,\mathrm{init}} = \bm{x}_{r,\mathrm{apex}}$ and $\bm{x}_{r,\mathrm{goal}} = \bm{x}_{r,\mathrm{apex}}$.To ensure a feasible transition between different juggling trajectories from the database, the robot's trajectories are designed to start and finish in the same joint state for all trajectories in the database. This setup is a prerequisite for implementing the periodic controller in Section~\ref{sec:controller}.

\subsection{Trajectory Transitioning via Time Reparameterization}\label{sec:controller}

As each control cycle starts and ends in the ball apex position, which is also associated with the robot's dynamic state that ensures a smooth transition between cycles, timing deviations require an additional correction strategy. Namely, the robot must reach its target state exactly when the ball reaches its apex. However, disturbances and model inaccuracies can cause the ball's actual trajectory to deviate from the planned trajectory, often resulting in a discrepancy between predicted and actual time point for reaching apex $t_{\mathrm{goal}}$. To prevent the resulting asynchrony, an adaptive mechanism is implemented to rescale the tool's trajectory timing, effectively ensuring that the tool can start the next trajectory as soon as the ball reaches its apex $t_{\text{goal}}$ . This online adaptation of the motion's temporal execution is enabled by a path-velocity decomposition method \cite{Lynch_Park_2017}. Given the optimal control solution as a discrete-time joint trajectory $\{\vec{q}_{\text{orig}}(t_i),\, \dot{\vec{q}}_{\text{orig}}(t_i)\}$, the solution is decomposed into a time-independent geometric path, $\vec{q}_{\text{path}}(s)$. This path is parameterized by its arc length $s \in [0,\, s_{\mathrm{total}}]$ and is represented by a cubic spline interpolated through the original joint configuration waypoints. During online execution, when the ball is launched at $t_{\mathrm{release}}$ and the apex time $t_{\mathrm{to\_apex}}$ is estimated, the trajectory's duration is reparameterized to a new end time $t_{\text{apex}}$. This is achieved by computing a new, smooth timing law, $s_{\mathrm{new}}(t)$, as the solution to a boundary value problem for the interval $[t_{\text{release}},\ t_{\text{apex}}]$. This ensures $C^1$-continuity at the transition point $t_{\mathrm{release}}$. The four boundary conditions are defined as:

\begin{figure}[t]
\vspace{0.17cm}
\centering
\includegraphics[scale=0.42]{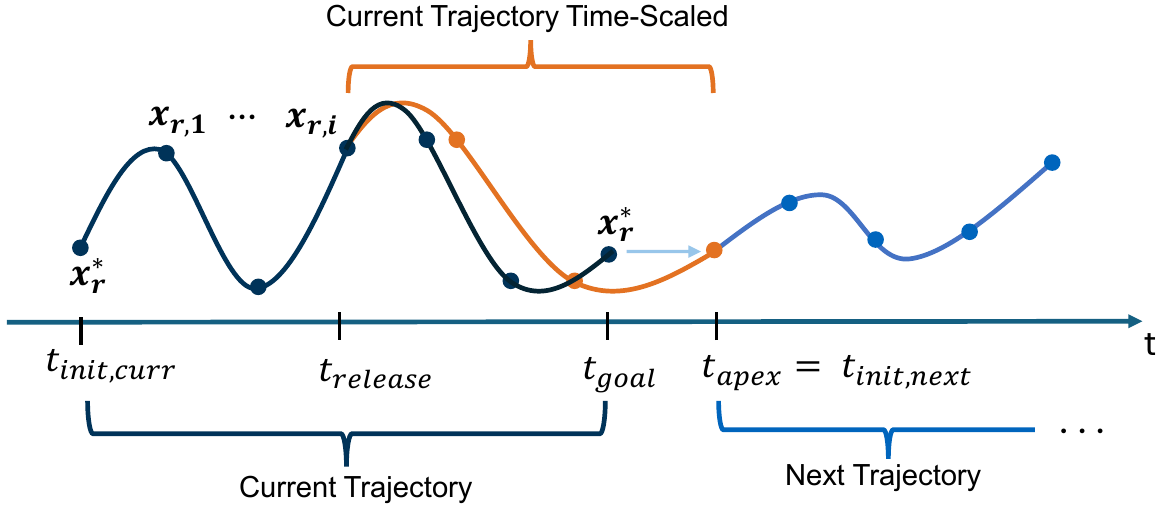}
\caption{Path-Preserving Trajectory Re-Timing. At \( t_{\mathrm{release}} \), the tool trajectory is re-scheduled on a new time scale, in case the calculated apex state of the tool (based on the ball state at detachment point) \( \bm{x}_{r,\mathrm{apex}} \) is different from expected one associated to the current tool trajectory. This also means that the current estimated time for reaching the apex \( t_{\mathrm{apex}} \) would be shifted from the originally planned one \( t_{\mathrm{goal}} \). Note that the new trajectory execution begins at \( t_{\mathrm{apex}} \).}
\label{fig:trajectory_scaling}
\vspace{-0.5cm}
\end{figure}

\begin{align*}
    s_{\text{new}}(t_{\text{release}}) &= s_{\text{orig}}(t_{\text{release}}), & s_{\text{new}}(t_{\text{apex}}) &= s_{\text{total}} \\
    \dot{s}_{\text{new}}(t_{\text{release}}) &= \| \dot{\vec{q}}_{\text{orig}}(t_{\text{release}}) \|, & \dot{s}_{\text{new}}(t_{\text{apex}}) &= \| \dot{\vec{q}}_{\text{start}} \|
\end{align*}
where $s_{\text{orig}}(t)$ is the timing law of the original trajectory and $\dot{\vec{q}}_{\text{start}}$ is the joint velocity component of the robot's apex state $\bm{x}_{r,\mathrm{apex}}$. This boundary value problem is solved using a piecewise cubic Hermite \cite{gautschi2011numerical} polynomial, which yields a unique and continuously differentiable solution for $s_{\text{new}}(t)$.

The new desired state for the robot is then synthesized from this new timing law. The desired joint position vector $\vec{q}_{\text{des}}(t)$ is obtained by evaluating the geometric path $\vec{q}_{\text{path}}$ at the newly timed arc length:
\begin{equation}
	\label{eq:joint_pos_from_path_clear}
	\vec{q}_{\text{des}}(t) = \vec{q}_{\text{path}}(s_{\text{new}}(t)).
\end{equation}
The corresponding desired joint velocity vector $\dot{\vec{q}}_{\text{des}}(t)$ is computed via the chain rule:
\begin{equation}
	\label{eq:joint_vel_from_path_clear}
	\dot{\vec{q}}_{\text{des}}(t) = \frac{d\vec{q}_{\text{path}}}{ds}(s_{\text{new}}(t)) \cdot \dot{s}_{\text{new}}(t).
\end{equation}
The resulting vectors $\vec{q}_{\text{des}}(t)$ and $\dot{\vec{q}}_{\text{des}}(t)$ serve as the reference signals for the low-level controller. This ensures that the original geometric path is followed while achieving a $C^1$-continuous transition between trajectory cycles.

\begin{figure*}[t]
\vspace*{0.2cm}
\centering
\includegraphics[scale=0.24]{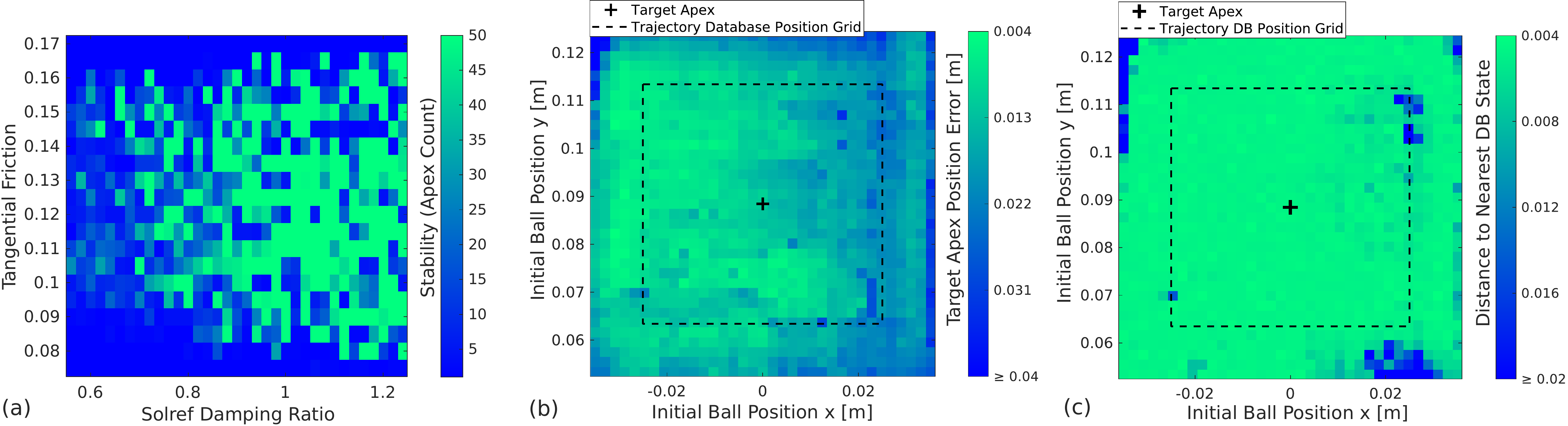}
\caption{Mapping the performance landscape of the juggling controller by analyzing the ball's resulting motion under variations in physical parameters and initial conditions. (a) Stability map of the control approach across different contact parameters. All parameters except for the tangential friction and damping ratio were kept at their MuJoCo default values. The color of each cell indicates the stability, measured by the maximum number of successful juggles (apex count) initiated from that parameter set; (b) Accuracy map of the worst-case positional error (Euclidean distance in x and y) between the achieved apex and the single nominal target apex. Each cell's value represents the maximum error found over 50 random trials initiated from that cell; (c) Robustness map of the worst-case state error (Euclidean distance in position and velocity) between the achieved apex and the nearest corresponding state in the pre-computed trajectory database.}
\label{fig:experiments}
\vspace{-0.5cm}
\end{figure*}

\section{Experiments} \label{sec:experiments}

\subsection{Model Validation}

An evaluation of the ball's dynamic response to the OCP-generated tool trajectories is performed in MuJoCo \cite{todorov2012mujoco} physics engine by simulating the direct interaction between the tool and the ball. Manipulator dynamics is omitted to isolate errors related to interaction behaviour only (e.g. ball deformation, rolling, etc.). Since our model operates in two dimensions, a tool was designed with a V-shaped groove to better align the simulated 3D system with the modelling assumptions by restricting the ball's movement.

\begin{figure*}[t]
\vspace{0.15cm}
\centering
\includegraphics[scale=0.39]{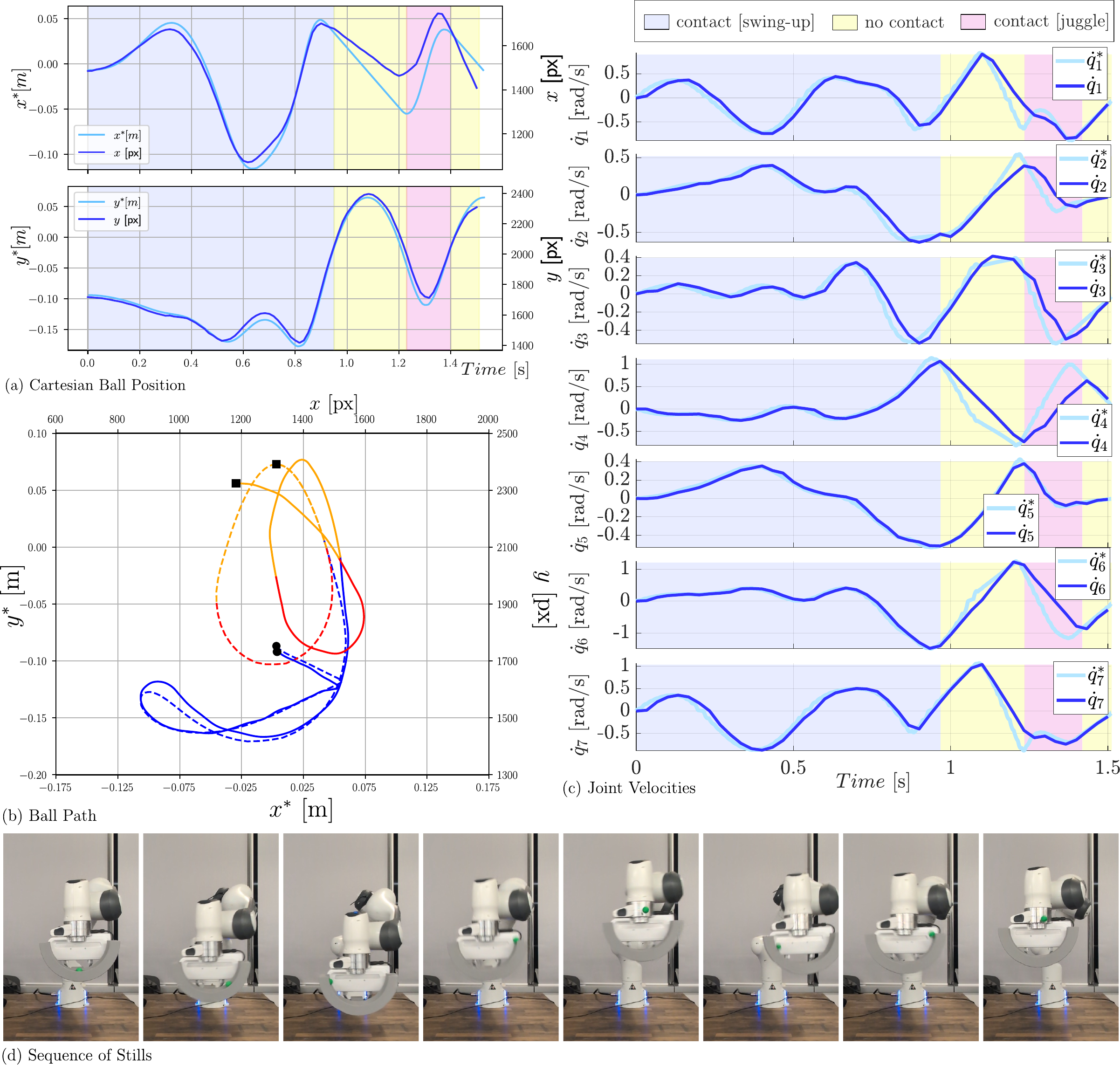}
\caption{Experiment performed on Panda Franka Emika Robot for swing up and one cycle of juggling using a velocity controller for trajectory tracking. The experimentally evaluated friction coefficient of $\mu=0.105$ is used for trajectory generation. a) and b) show the ball position in Cartesian space. The actual ball trajectory is extracted from a video recording. c) shows the tracking performance of the joint-level velocity controllers; and d) shows a sequence of stills of swing up and once cycle of juggling.}
\label{fig:hardware_experiments}
\vspace{-0.5cm}
\end{figure*}

As shown in Fig.~\ref{fig:Opt_vs_Sim}, the best match between the optimal and simulated trajectories is achieved when the simulation's friction coefficient is set lower than the model's. This suggests that the friction coefficient in our model acts as a lumped parameter, compensating for multiple energy dissipation phenomena present in the more realistic simulation. These phenomena include, for example, the increased effective friction from the V-shaped tool's double contact points and energy losses due to ball rolling.

\subsection{Controller Performance Analysis}

The controller's sensitivity to the contact model was analyzed via a parameter sweep across a range of tangential friction coefficients and solref damping ratios (which determines the elasticity of the impact), with the results presented in Fig.~\ref{fig:experiments}a. Each simulation run began with a standardized swing-up maneuver, at which point the closed-loop juggling controller took over. The stability for each parameter configuration was then measured by the total number of consecutive juggles (apex count) the system could sustain before the ball was dropped.

Using a stable set of contact parameters identified in the previous analysis, we then performed a grid-based Monte Carlo simulation to quantitatively evaluate the controller's robustness. A 7x7 cm grid of initial apex positions, centered around the nominal target state, was discretized into 2x2 mm cells. Within each cell, 50 initial ball states were randomly sampled from a uniform distribution, and horizontal velocities were sampled from a dynamic range based on the cell's x-position. For each sampled initial state, the decision policy selected the optimal recovery trajectory from the database, and a single juggle cycle was simulated in MuJoCo. From the recorded final apex states, the worst-case performance of each cell was evaluated and visualized in terms of target deviation (Fig.~\ref{fig:experiments}b) and from the point of view of robustness by the ability to choose a reasonably close database point with a recovery trajectory in Fig.~\ref{fig:experiments}c.

\subsection{Hardware Experiments}
For the hardware experiments, the circular tool was 3D-printed and attached to the robot's gripper. To calibrate our contact model, an initial system identification step was performed. We executed a series of swing-up trajectories and used external camera recordings to measure the resulting flight path of the ball, allowing us to empirically determine the effective friction coefficient of the physical tool-ball pair. With this calibrated friction parameter, a trajectory consisting of a swing-up and one subsequent juggling maneuver was generated. This trajectory was then mapped to the joint space and executed on the robot. A stop-motion of the experiment, data on controller performance, alongside the resulting ball trajectory, can be seen in Fig.~\ref{fig:hardware_experiments}. Experiments are limited to a swing-up and single juggle cycle for evaluating the presented model-based framework and robot performance abilities.

\section{Conclusion}

Highly dynamic and timing-sensitive systems such as non-prehensile robot juggling require an online strategy to counteract ball deviations and maintain the desired behavior. For that, a model-based two-stage optimal control framework has been presented. In the first stage, the \emph{{Task-Space OCP}} generates juggling motions from an analytical 2D model. In the second stage, \emph{{Joint-Space OCP}} takes the resulting 2D trajectories and maps them to dynamically feasible joint-space trajectories. This process is repeatedly solved offline to generate a library of pre-computed correction trajectories for the predicted perturbed apex state range. A decision policy is then deployed online. It takes the ball state and determines the correction trajectory to be executed next. To ensure a smooth transition between different DB trajectories, a trajectory re-parametrisation approach has been adopted. Performance was evaluated in both simulation and hardware experiments using the Franka Panda robot with a circular tool and a ball to perform swing-up and juggling tasks.

\section*{Acknowledgement}
The authors would like to thank Bingkun Huang for fruitful discussions and help with the hardware experiments. This work has partially been supported by the German Federal Ministry of Research, Technology and Space (BMFTR) under the Robotics Institute Germany (RIG), grant no. 16ME0997K.



\begin{thebibliography}{10}
\providecommand{\url}[1]{#1}
\csname url@samestyle\endcsname
\providecommand{\newblock}{\relax}
\providecommand{\bibinfo}[2]{#2}
\providecommand{\BIBentrySTDinterwordspacing}{\spaceskip=0pt\relax}
\providecommand{\BIBentryALTinterwordstretchfactor}{4}
\providecommand{\BIBentryALTinterwordspacing}{\spaceskip=\fontdimen2\font plus
\BIBentryALTinterwordstretchfactor\fontdimen3\font minus
  \fontdimen4\font\relax}
\providecommand{\BIBforeignlanguage}[2]{{%
\expandafter\ifx\csname l@#1\endcsname\relax
\typeout{** WARNING: IEEEtran.bst: No hyphenation pattern has been}%
\typeout{** loaded for the language `#1'. Using the pattern for}%
\typeout{** the default language instead.}%
\else
\language=\csname l@#1\endcsname
\fi
#2}}
\providecommand{\BIBdecl}{\relax}
\BIBdecl

\bibitem{ruggiero2018nonprehensile}
F.~Ruggiero, V.~Lippiello, and B.~Siciliano, ``Nonprehensile dynamic
  manipulation: A survey,'' \emph{IEEE Robotics and Automation Letters},
  vol.~3, no.~3, pp. 1711--1718, 2018.

\bibitem{muchacho2022solution}
R.~I.~C. Muchacho, R.~Laha, L.~F. Figueredo, and S.~Haddadin, ``A solution to
  slosh-free robot trajectory optimization,'' in \emph{2022 IEEE/RSJ
  International Conference on Intelligent Robots and Systems (IROS)}.

\bibitem{buehler1987robotics}
M.~Buehler and D.~E. Koditschek, ``Robotics in an intermittent dynamical
  environment: A prelude to juggling,'' 1987.

\bibitem{aboaf1988task}
E.~W. Aboaf, C.~G. Atkeson, and D.~J. Reinkensmeyer, ``Task-level robot
  learning,'' in \emph{Proceedings. 1988 IEEE International Conference on
  Robotics and Automation}.\hskip 1em plus 0.5em minus 0.4em\relax IEEE, 1988,
  pp. 1309--1310.

\bibitem{schaal1996one}
S.~Schaal, C.~G. Atkeson, and D.~Sternad, ``One-handed juggling: A dynamical
  approach to a rhythmic movement task,'' \emph{Journal of Motor Behavior},
  vol.~28, no.~2, pp. 165--183, 1996.

\bibitem{schaal1993open}
S.~Schaal and C.~G. Atkeson, ``Open loop stable control strategies for robot
  juggling,'' in \emph{Proceedings IEEE International Conference on Robotics
  and Automation}.\hskip 1em plus 0.5em minus 0.4em\relax IEEE, 1993, pp.
  913--918.

\bibitem{reist2012design}
P.~Reist and R.~D'Andrea, ``Design and analysis of a blind juggling robot,''
  \emph{IEEE Transactions on Robotics}, vol.~28, no.~6, pp. 1228--1243, 2012.

\bibitem{beek1995science}
P.~J. Beek and A.~Lewbel, ``The science of juggling,'' \emph{Scientific
  American}, vol. 273, no.~5, pp. 92--97, 1995.

\bibitem{polster2003mathematics}
B.~Polster, \emph{The mathematics of juggling}, 2003.

\bibitem{lampariello2011trajectory}
R.~Lampariello, D.~Nguyen-Tuong, C.~Castellini, G.~Hirzinger, and J.~Peters,
  ``Trajectory planning for optimal robot catching in real-time,'' in
  \emph{IEEE International Conference on Robotics and Automation}, 2011, pp.
  3719--3726.

\bibitem{bauml2010kinematically}
B.~B{\"a}uml, T.~Wimb{\"o}ck, and G.~Hirzinger, ``Kinematically optimal
  catching a flying ball with a hand-arm-system,'' in \emph{IEEE/RSJ
  International Conference on Intelligent Robots and Systems}, 2010, pp.
  2592--2599.

\bibitem{hove1991experiments}
B.~Hove and J.-J.~E. Slotine, ``Experiments in robotic catching,'' in
  \emph{1991 American Control Conference}.\hskip 1em plus 0.5em minus
  0.4em\relax IEEE, 1991, pp. 380--386.

\bibitem{namiki2014ball}
A.~Namiki and N.~Itoi, ``Ball catching in kendama game by estimating grasp
  conditions based on a high-speed vision system and tactile sensors,'' in
  \emph{2014 IEEE-RAS International Conference on Humanoid Robots}.\hskip 1em
  plus 0.5em minus 0.4em\relax IEEE, 2014, pp. 634--639.

\bibitem{kim2014catching}
S.~Kim, A.~Shukla, and A.~Billard, ``Catching objects in flight,'' \emph{IEEE
  Transactions on Robotics}, vol.~30, no.~5, pp. 1049--1065, 2014.

\bibitem{montana1988kinematics}
D.~J. Montana, ``The kinematics of contact and grasp,'' \emph{The International
  Journal of Robotics Research}, vol.~7, no.~3, pp. 17--32, 1988.

\bibitem{woodruff2020motion}
J.~Z. Woodruff, S.~Ren, and K.~M. Lynch, ``Motion planning and feedback control
  of rolling bodies,'' \emph{IEEE Access}, vol.~8, 2020.

\bibitem{woodruff2023robotic}
J.~Z. Woodruff and K.~M. Lynch, ``Robotic contact juggling,'' \emph{IEEE
  Transactions on Robotics}, vol.~39, no.~3, pp. 1964--1981, 2023.

\bibitem{lagrange1853mecanique}
J.~L. Lagrange, \emph{M{\'e}canique analytique}.\hskip 1em plus 0.5em minus
  0.4em\relax Mallet-Bachelier, 1853, vol.~1.

\bibitem{posa2014direct}
M.~Posa, C.~Cantu, and R.~Tedrake, ``A direct method for trajectory
  optimization of rigid bodies through contact,'' \emph{The International
  Journal of Robotics Research}, vol.~33, no.~1, pp. 69--81, 2014.

\bibitem{yao2025synthesis}
H.~Yao, R.~Laha, A.~Sinha, J.~Hall, L.~F. Figueredo, N.~Chakraborty, and
  S.~Haddadin, ``On the synthesis of reactive collision-free whole-body robot
  motions: A complementarity-based approach,'' in \emph{2025 IEEE International
  Conference on Robotics and Automation (ICRA)}.

\bibitem{andersson2019casadi}
J.~A. Andersson, J.~Gillis, G.~Horn, J.~B. Rawlings, and M.~Diehl, ``Casadi: a
  software framework for nonlinear optimization and optimal control,''
  \emph{Mathematical Programming Computation}, vol.~11, 2019.

\bibitem{wachter2006implementation}
A.~W{\"a}chter and L.~T. Biegler, ``On the implementation of an interior-point
  filter line-search algorithm for large-scale nonlinear programming,''
  \emph{Mathematical programming}, vol. 106, pp. 25--57, 2006.

\bibitem{hsl2007collection}
A.~HSL, ``Collection of fortran codes for large scale scientific computation,''
  \emph{See http://www. hsl. rl. ac. uk}, 2007.

\bibitem{haddadin2022franka}
S.~Haddadin, S.~Parusel, L.~Johannsmeier, S.~Golz, S.~Gabl, F.~Walch,
  M.~Sabaghian, C.~J{\"a}hne, L.~Hausperger, and S.~Haddadin, ``The franka
  emika robot: A reference platform for robotics research and education,''
  \emph{IEEE Robotics \& Automation Magazine}, vol.~29, no.~2, pp. 46--64,
  2022.

\bibitem{carpentier2019pinocchio}
J.~Carpentier, G.~Saurel, G.~Buondonno \emph{et~al.}, ``The pinocchio c++
  library -- a fast and flexible implementation of rigid body dynamics
  algorithms and their analytical derivatives,'' in \emph{IEEE International
  Symposium on System Integrations (SII)}, 2019.

\bibitem{Ramadani2025}
\BIBentryALTinterwordspacing
J.~K. Ramadani, ``\BIBforeignlanguage{en}{Robust optimal control approach for
  stable ball juggling using a 7-dof manipulator},'' Master's thesis, Technical
  University of Munich, 2025. [Online]. Available:
  \url{https://mediatum.ub.tum.de/1843356}
\BIBentrySTDinterwordspacing

\bibitem{Lynch_Park_2017}
K.~M. Lynch and F.~C. Park, \emph{Modern Robotics: Mechanics, Planning, and
  Control}.\hskip 1em plus 0.5em minus 0.4em\relax Cambridge University Press,
  2017.

\bibitem{gautschi2011numerical}
W.~Gautschi, \emph{Numerical analysis}.\hskip 1em plus 0.5em minus 0.4em\relax
  Springer Science \& Business Media, 2011.

\bibitem{todorov2012mujoco}
E.~Todorov, T.~Erez, and Y.~Tassa, ``Mujoco: A physics engine for model-based
  control,'' in \emph{2012 IEEE/RSJ International Conference on Intelligent
  Robots and Systems}.\hskip 1em plus 0.5em minus 0.4em\relax IEEE, 2012, pp.
  5026--5033.

\end{thebibliography}
\end{document}